\pdfoutput=1

\documentclass[11pt]{article}

\usepackage[final]{acl}

\usepackage{times}
\usepackage{latexsym}
\usepackage{amsmath,amssymb,amsfonts}%
\usepackage[T1]{fontenc}

\usepackage[utf8]{inputenc}

\usepackage{microtype}
\usepackage{array}
\usepackage{multirow}%
\usepackage{xcolor}
\usepackage{colortbl} 
\usepackage{inconsolata}
\usepackage{booktabs}%
\usepackage{subcaption}

\usepackage{graphicx}

\title{Reference-free Hallucination Detection for \\  Large Vision-Language Models}
\definecolor{lightgray}{gray}{0.9}
\definecolor{lightred}{rgb}{0.98, 0.81, 0.82}
\definecolor{lightgreen}{rgb}{0.88, 1.0, 0.78}
\definecolor{lightyellow}{rgb}{1.0, 1.0, 0.68}

\author{
${\textbf{Qing Li}}^{1*}$,
${\textbf{Jiahui Geng}}^{1*}$,
${\textbf{Chenyang Lyu}}^1$,
${\textbf{Derui Zhu}}^2$, 
${\textbf{Maxim Panov}}^1$,
${\textbf{Fakhri Karray}}^{1\dagger}$
\\
$^1$ Mohamed bin Zayed University of Artificial Intelligence \\
\{qing.li, jiahui.geng, chenyang.lyu, maxim.panov, fakhri.karray\}@mbzuai.ac.ae \\
$^2$ Technical University of Munich \\
derui.zhu@tum.de
}

\begin{document}
\maketitle

\begin{abstract}

Large vision-language models (LVLMs) have made significant progress in recent years. While LVLMs exhibit excellent ability in language understanding, question answering, and conversations of visual inputs, they are prone to producing hallucinations. While several methods are proposed to evaluate the hallucinations in LVLMs, most are reference-based and depend on external tools, which complicates their practical application. To assess the viability of alternative methods, it is critical to understand whether the reference-free approaches, which do not rely on any external tools, can efficiently detect hallucinations. Therefore, we initiate an exploratory study to demonstrate the effectiveness of different reference-free solutions in detecting hallucinations in LVLMs. In particular, we conduct an extensive study on three kinds of techniques: uncertainty-based, consistency-based, and supervised uncertainty quantification methods on four representative LVLMs across two different tasks. The empirical results show that the reference-free approaches are capable of effectively detecting non-factual responses in LVLMs, with the supervised uncertainty quantification method outperforming the others, achieving the best performance across different settings.

\footnote{*Equal contribution.}
\footnote{\textdagger Corresponding author.}
 
\end{abstract}

\section{Introduction}

\begin{figure}[t!]
    \centering
    \includegraphics[width=0.95\linewidth]{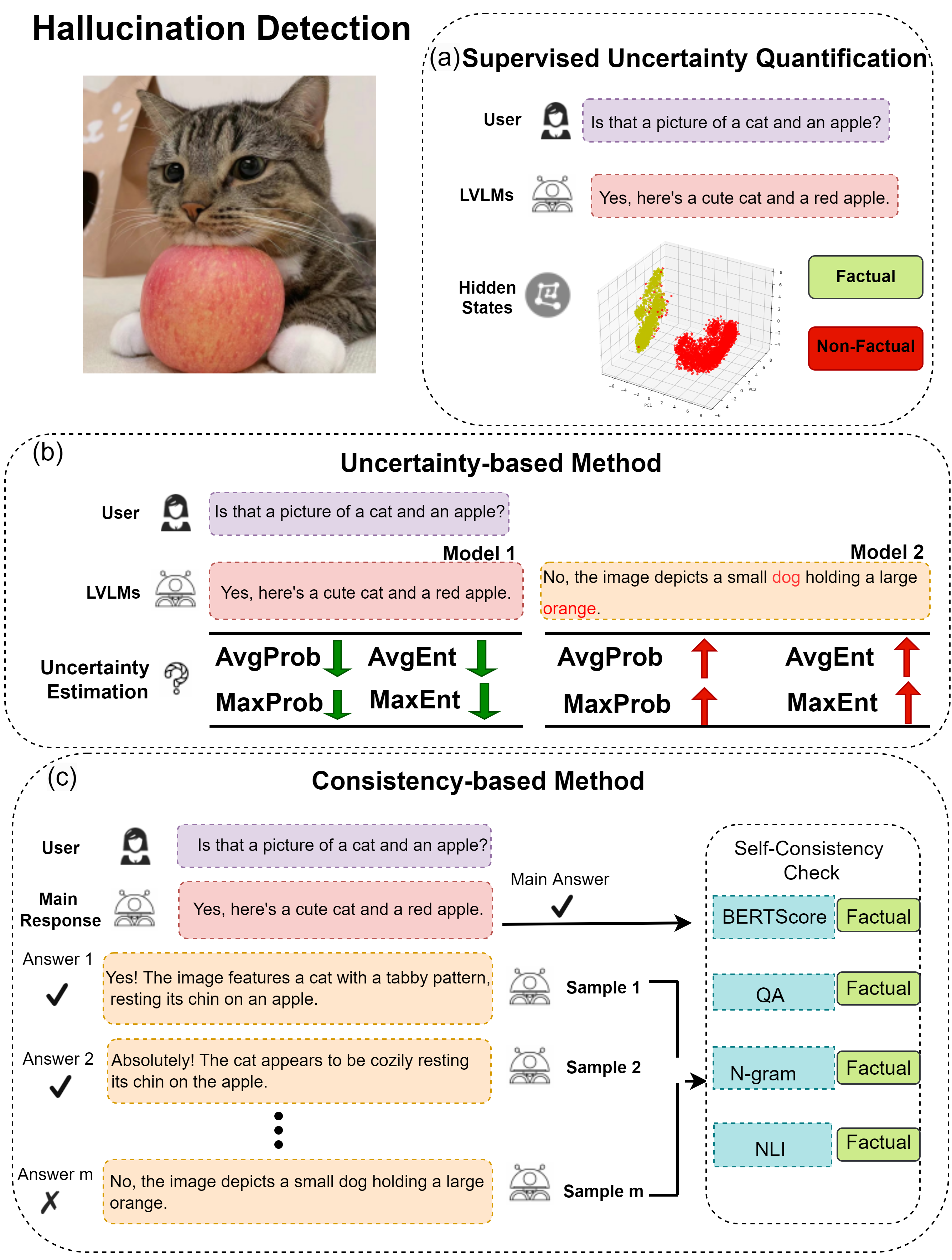}
    \caption{Reference-free Hallucination Detection Methods used in this work.}
    \label{fig:all:methods}
\end{figure}

Large vision-language models (LVLMs) like LLaVA~\cite{liu2023improvedllava}, MiniGPT-4~\cite{zhu2023minigpt} have demonstrated remarkable capabilities in understanding and generating complex visual and textual content. However, an emerging concern with these models is their tendency towards hallucination~\cite{zhou2023analyzing,zhu-etal-2024-pollmgraph,geng2024multimodal}. For instance, a model might describe an object or event in the generated text that is not present in the input image. This phenomenon poses challenges for the reliability of LVLMs, highlighting intrinsic limitations of current models in maintaining coherence between visual inputs and textual descriptions~\cite{huang2023opera}. 

Most existing methods of assessing hallucinations rely on external models. \citet{li2023evaluating} propose POPE, which requires automated segmentation tools like SEEM~\cite{zou2024segment} to identify present and absent objects in images. Faithscore~\cite{jing2023faithscore} utilizes a visual entailment model~\cite{xie2019visual} to validate the factuality, focusing on object attributes and their relationships. Nonetheless, these approaches increase the computational cost and are restricted by the capabilities of these models. 

\begin{figure}[t!]
    \centering
    \includegraphics[width=\linewidth]{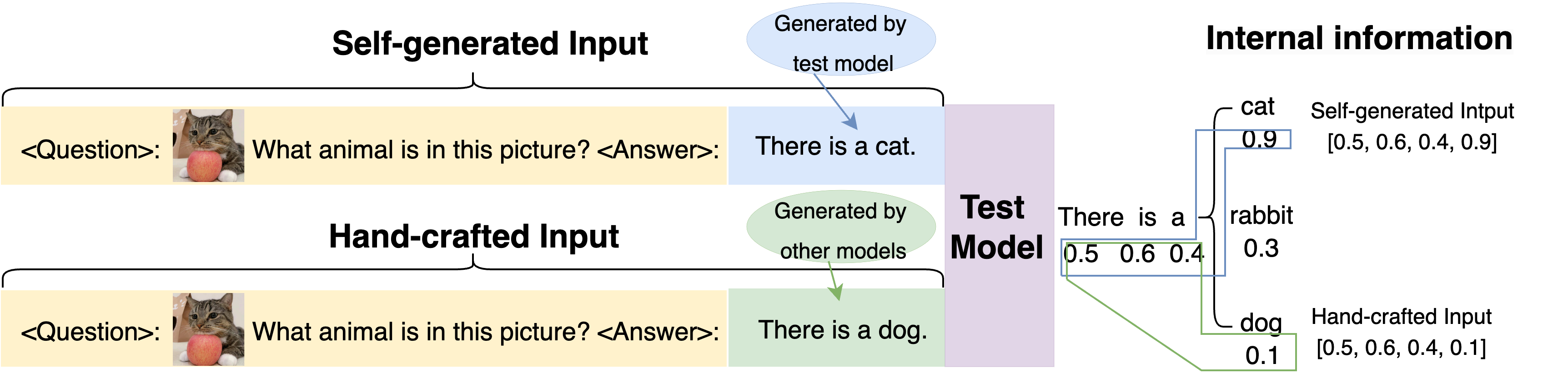}
    \caption{Two data sources: self-generated and hand-crafted data. The test model assigns a probability of 0.9 to ``cat'' in self-generated input, while in hand-crafted input, ``dog'' receives a probability of 0.1.}
    \label{fig:promt_response.png}
\end{figure}

Reference-free methods are a promising approach to address this issue, primarily utilizing the model's own knowledge. They have attracted great research interest in Large Language Models (LLMs), with typical methods including the uncertainty-based methods~\cite{hu2023uncertainty,geng2023survey, huang2023look,vazhentsev-etal-2023-hybrid,DBLP:conf/emnlp/FadeevaVTVPFVGP23}, consistency-based methods~\cite{manakul2023selfcheckgpt} and supervised uncertainty quantification (SUQ) approach~\cite{chen2023beyond,azaria2023internal}. However, their potential in the context of LVLMs is unknown. They have not yet been systematically studied, possibly due to the complexity of multimodal data. We are committed to bridging this gap.

In this work, we collect and implement a battery of state-of-the-art methods, including four metrics of uncertainty-based methods, four variants of consistency-based methods, and the supervised uncertainty quantification method. The experiments are conducted on five LVLMs with different types, versions, and sizes on Yes-and-No and Open-ended tasks. The main contributions of this paper are summarized as follows:

\begin{enumerate}
    \item We comprehensively measure the performance of different reference-free approaches in detecting hallucination. 

    \item We demonstrate that the supervised uncertainty quantification (SUQ) method performs best across various settings. 

    \item We contribute the Image-Hallucination Annotation Dataset (IHAD), a manually annotated, sentence-level dataset created by prompting LLaVA-v1.5-7b.
\end{enumerate}

\section{Reference-free Hallucination Detection Methods}

\begin{table}[t!]
\small
\centering
\begin{tabular}{m{0.05cm}<{\centering} m{1.5cm}<{\raggedright} m{0.8cm}<{\centering} m{0.8cm}<{\centering} m{0.8cm}<{\centering} m{0.8cm}<{\centering}}
\toprule
&  Method  & Mini-GPT-4v     & LLaVA-v1.5-7b         &LLaVA-v1.6-vicuna-7b     &LLaVA-v1.6-mistral-7b  \\
\midrule
\multirow{5}{*}{\begin{tabular}[c]{@{}l@{}} \rotatebox{90}{Adv} \end{tabular}} 
&Random & 48.73     & 48.73   & 48.73     & 48.73           \\ 
& \multicolumn{5}{l}{\cellcolor{lightgray}{Uncertainty-based Methods}}  \\
& \text{AvgProb}  & \colorbox{lightred}{\parbox{0.04\textwidth}{77.43}} & \colorbox{lightred}{\parbox{0.04\textwidth}{86.82}}   & \colorbox{lightred}{\parbox{0.04\textwidth}{91.80}}     &  \colorbox{lightred}{\parbox{0.04\textwidth}{92.04}}    \\
& \text{AvgEnt}      & 48.89 & 50.05  & 48.90     & 48.92       \\
& \multicolumn{5}{l}{\cellcolor{lightgray}{Supervised Uncertainty Quantification Method}} \\
&SUQ  & \textbf{87.23}  & \textbf{92.35}   & \textbf{93.32}    & \textbf{93.50}             \\                                                                           
\midrule
\multirow{5}{*}{\begin{tabular}[c]{@{}l@{}} \rotatebox{90}{Ran} \end{tabular}}   
&Baseline    & 48.73     & 48.73   &   48.73 & 48.73            \\
& \multicolumn{5}{l}{\cellcolor{lightgray}{Uncertainty-based Methods}}  \\
& \text{AvgProb}    & \colorbox{lightred}{\parbox{0.04\textwidth}{86.62}} &  \colorbox{lightred}{\parbox{0.04\textwidth}{95.89}}   & \colorbox{lightred}{\parbox{0.04\textwidth}{96.23}}   & \colorbox{lightred}{\parbox{0.04\textwidth}{96.99}}    \\
& \text{AvgEnt}   & 48.76  &  50.25  & 49.91   & 50.41       \\
& \multicolumn{5}{l}{\cellcolor{lightgray}{Supervised Uncertainty Quantification Method}} \\
&SUQ   &  \textbf{94.44} & \textbf{97.86} & \textbf{98.68}  &  \textbf{98.33}               \\                                                                           
\midrule
\multirow{5}{*}{\begin{tabular}[c]{@{}l@{}} \rotatebox{90}{Pop} \end{tabular}} 
&Baseline    & 48.73     & 48.73  &  48.73  & 48.73              \\
& \multicolumn{5}{l}{\cellcolor{lightgray}{Uncertainty-based Methods}}   \\
& \text{AvgProb}    & \colorbox{lightred}{\parbox{0.04\textwidth}{82.11}} & \colorbox{lightred}{\parbox{0.04\textwidth}{92.73}}  & \colorbox{lightred}{\parbox{0.04\textwidth}{93.57}}    &  \colorbox{lightred}{\parbox{0.04\textwidth}{95.06}}  \\
& \text{AvgEnt}  & 48.50 & 49.76  & 50.63   & 51.39  \\
& \multicolumn{5}{l}{\cellcolor{lightgray}{Supervised Uncertainty Quantification Method}} \\
&SUQ    & \textbf{91.02}   & \textbf{97.85} & \textbf{98.11}  & \textbf{98.50}      \\                                                                            
\midrule
\multirow{5}{*}{\begin{tabular}[c]{@{}l@{}} \rotatebox{90}{GQA} \end{tabular}}
&Baseline    & 48.47     & 48.47   &  48.47  & 48.47         \\
& \multicolumn{5}{l}{\cellcolor{lightgray}{Uncertainty-based Methods}}  \\
& \text{AvgProb} & \colorbox{lightred}{\parbox{0.04\textwidth}{66.87}}  & \colorbox{lightred}{\parbox{0.04\textwidth}{78.45}}  & \colorbox{lightred}{\parbox{0.04\textwidth}{80.31}}  &  \colorbox{lightred}{\parbox{0.04\textwidth}{79.15}}  \\
& \text{AvgEnt} & 48.75 & 49.88  & 48.95    &  49.33  \\
& \multicolumn{5}{l}{\cellcolor{lightgray}{Supervised Uncertainty Quantification Method}} \\
&SUQ    & \textbf{69.14}    & \textbf{82.84} & \textbf{84.57} & \textbf{83.25} \\
\bottomrule
\end{tabular}
\caption{AUC-PR for Yes-or-No tasks.}
\label{tb:binary:output:tasks}
\end{table}

\noindent\textbf{Uncertainty-based methods} have the hypothesis that when LVLMs are uncertain about generated information, generated tokens often have higher uncertainty, shown in Figure~\ref{fig:all:methods}{\color{darkblue}(b)}. Uncertainty-based methods are extensively studied for hallucination detection in LLMs~\cite{hu2023uncertainty, geng2023survey, huang2023look,fadeeva2024factchecking}, and are also used as benchmark techniques~\cite{zhu-etal-2024-pollmgraph, manakul2023selfcheckgpt}.
We adopt the four proposed metrics: \text{AvgProb}, \text{AvgEnt}, \text{MaxProb}, \text{MaxEnt} to aggregate token-level uncertainty to measure sentence-level and passage-level uncertainty. Appendix~\ref{sec:appendix:uncertainty-based-methods} provides detailed descriptions of these metrics.

\noindent\textbf{Consistency-based methods} mainly include self-consistency, cross-question consistency, and cross-model consistency~\cite{zhang2023sac}. In this work, we focus on four variants -- BERTScore, Question Answering (QA), Unigram, Natural Language Inference (NLI) -- of the self-consistency method proposed in~\cite{manakul2023selfcheckgpt} to detect non-factual information for LVLMs, as demonstrated in Figure~\ref{fig:all:methods}{\color{darkblue}(c)}. The hallucination score for the \(i\)-th sentence, denoted by $\mathcal{S}(i)$, ranges from 0.0 to 1.0, where a score close to 0.0 indicates the sentence is accurate, and a score near 1.0 suggests the sentence is hallucinated. More detailed information can be found in Appendix~\ref{sec:appendix:consistency-based-methods}.

\noindent\textbf{Supervised uncertainty quantification (SUQ).} To analyze and understand these internal states, SUQ method~\cite{chen2023beyond,azaria2023internal} shown in Figure~\ref{fig:all:methods}{\color{darkblue}(a)} train a classifier, referred to as a \emph{probe}, on a dataset containing labeled examples. This classifier outputs the likelihood of a statement being truthful by analyzing the hidden layer activations of the LVLM during the reading or generation of states.

\begin{table}[t]
\small
\centering
\begin{tabular}{m{0.04cm}<{\raggedright} m{1.2cm}<{\raggedright} m{0.7cm}<{\centering} m{0.7cm}<{\centering} m{0.7cm}<{\centering} m{0.7cm}<{\centering} m{0.7cm}<{\centering}}
\toprule
 &  Method  & Mini-GPT-4v     & LLaVA-v1.5-7b         &LLaVA-v1.6-vicuna-7b       &LLaVA-v1.6-mistral-7b  &LLaVA-v1.6-vicuna-13b \\
\midrule
\multirow{14}{*}{\begin{tabular}[c]{@{}l@{}} \rotatebox{90}{M-Hal} \end{tabular}} 
&Baseline & 29.29     & 29.29   & 29.29     & 29.29     & 29.29       \\ 
& \multicolumn{6}{l}{\cellcolor{lightgray}{Uncertainty-based Methods}}  \\
& \text{AvgProb}  & 30.52  & \colorbox{lightred}{\parbox{0.04\textwidth}{37.22}}  & 38.61  & 37.54  & \colorbox{lightred}{\parbox{0.04\textwidth}{43.30}}\\
& \text{AvgEnt} & 29.92  &32.18  & 39.13 & 37.22 & 40.11 \\
& \text{MaxProb} & \colorbox{lightred}{\parbox{0.04\textwidth}{33.45}} &32.05  & 34.86 &34.73 & 40.03 \\
& \text{MaxEnt} & 29.96 & 29.71 & \colorbox{lightred}{\parbox{0.04\textwidth}{39.71}} & \colorbox{lightred}{\parbox{0.04\textwidth}{39.62}} & 33.14 \\
& \multicolumn{6}{l}{\cellcolor{lightgray}{Consistency-based Methods}}  \\
& QA          & 35.08  &40.27 & 41.44  & 39.48 & 46.38 \\
& BERTScore   & 37.21  &39.13 & 51.16  & 51.64 & 54.12 \\
& Unigram (max)    & 37.63  & 38.68  & 39.62 & 40.08 & 49.23 \\
& NLI         & \colorbox{lightyellow}{\parbox{0.04\textwidth}{42.14}}&   \colorbox{lightyellow}{\parbox{0.04\textwidth}{44.09}} &\colorbox{lightyellow}{\parbox{0.04\textwidth}{52.39}} & \colorbox{lightyellow}{\parbox{0.04\textwidth}{51.91}} & \colorbox{lightyellow}{\parbox{0.04\textwidth}{57.36}} \\ 
& \multicolumn{6}{l}{\cellcolor{lightgray}{Supervised Uncertainty Quantification Method}} \\
&SUQ  & \textbf{62.53} & \textbf{61.23} & \textbf{65.26} & \textbf{64.78}   & \textbf{70.12}  \\                                                                           
\midrule
\multirow{14}{*}{\begin{tabular}[c]{@{}l@{}} \rotatebox{90}{IHAD} \end{tabular}}   
&Baseline    & 35.71     & 35.71   &   35.71 & 35.71    & 35.71        \\
& \multicolumn{6}{l}{\cellcolor{lightgray}{Uncertainty-based Methods}}  \\
& \text{AvgProb} & \colorbox{lightred}{\parbox{0.04\textwidth}{36.16}} & \colorbox{lightred}{\parbox{0.04\textwidth}{42.35}} & \colorbox{lightred}{\parbox{0.04\textwidth}{50.88}} & \colorbox{lightred}{\parbox{0.04\textwidth}{50.75}}  & \colorbox{lightred}{\parbox{0.04\textwidth}{53.42}}  \\
& \text{AvgEnt}  &35.81  & 37.18  &48.38   & 46.86   & 50.04  \\
& \text{MaxProb}  & 35.97 &41.89  &46.24  & 45.96  & 45.21 \\
& \text{MaxEnt} & 35.53 &35.88  &50.85  & 46.65  & 50.33  \\ 
& \multicolumn{6}{l}{\cellcolor{lightgray}{Consistency-based Methods}}  \\
& QA    & 41.34  & \colorbox{lightyellow}{\parbox{0.04\textwidth}{54.72}} & 48.15&  49.19  & 58.12\\
& BERTScore & 40.29 & 42.37&  49.03 & 47.87  & 50.33 \\
& Unigram (max)  &39.26  & 39.06 & 43.13 & 44.10  & 49.79\\
& NLI  & \colorbox{lightyellow}{\parbox{0.04\textwidth}{49.59}}  & 50.80 &\colorbox{lightyellow}{\parbox{0.04\textwidth}{50.94}}  &\colorbox{lightyellow}{\parbox{0.04\textwidth}{58.41}} & \colorbox{lightyellow}{\parbox{0.04\textwidth}{62.58}}\\
& \multicolumn{6}{l}{\cellcolor{lightgray}{Supervised Uncertainty Quantification Method}} \\
&SUQ  & \textbf{62.73}  & \textbf{63.85}  & \textbf{65.74}  & \textbf{64.17} & \textbf{69.19}\\     \bottomrule
\end{tabular}
\caption{AUC-PR for sentence-level Open-ended tasks.}
\label{tb:long:form:tasks:sentence}
\end{table}

\begin{table}[t]
\small
\centering
\begin{tabular}{m{0.04cm}<{\raggedright} m{1.2cm}<{\raggedright} m{0.7cm}<{\centering} m{0.7cm}<{\centering} m{0.7cm}<{\centering} m{0.7cm}<{\centering} m{0.7cm}<{\centering}}
\toprule
 &  Method  & Mini-GPT-4v     & LLaVA-v1.5-7b         &LLaVA-v1.6-vicuna-7b       &LLaVA-v1.6-mistral-7b & LLaVA-v1.6-vicuna-13b \\
\midrule
\multirow{8}{*}{\begin{tabular}[c]{@{}l@{}} \rotatebox{90}{ M-Hal} \end{tabular}} 
&Baseline & 64.05     & 64.05   & 64.05     & 64.05 & 64.05 \\ 
& \multicolumn{6}{l}{\cellcolor{lightgray}{Uncertainty-based Methods}}  \\
& \text{AvgProb}  & \colorbox{lightred}{\parbox{0.04\textwidth}{70.28}} & \colorbox{lightred}{\parbox{0.04\textwidth}{72.78}}  & 72.89 & 73.32 & \colorbox{lightred}{\parbox{0.04\textwidth}{81.23}} \\
& \text{AvgEnt} & 64.89 &64.44  & 70.83  &70.05  & 72.97\\
& \text{MaxProb} & 66.68 &  68.97  &72.67  & 70.37 & 73.56 \\
& \text{MaxEnt} & 68.75 & 64.99 & \colorbox{lightred}{\parbox{0.04\textwidth}{77.17}}  & \colorbox{lightred}{\parbox{0.04\textwidth}{78.62}} & 80.23 \\
& \multicolumn{6}{l}{\cellcolor{lightgray}{Supervised Uncertainty Quantification Method}} \\
&SUQ  & \textbf{84.92} & \textbf{89.41} & \textbf{85.91}& \textbf{86.54} & \textbf{91.22} \\  \midrule
\multirow{8}{*}{\begin{tabular}[c]{@{}l@{}} \rotatebox{90}{IHAD} \end{tabular}}   
& Baseline  & 83.00 & 83.00  & 83.00  & 83.00 & 83.00 \\
& \multicolumn{6}{l}{\cellcolor{lightgray}{Uncertainty-based Methods}}  \\
& \text{AvgProb} & \colorbox{lightred}{\parbox{0.04\textwidth}{84.29}} & 87.50 & 87.38  & 87.63 & 90.23 \\
& \text{AvgEnt} &  83.06  & 85.64 & 88.41 & 87.45  & 89.41    \\
& \text{MaxProb}  & 84.17 & \colorbox{lightred}{\parbox{0.04\textwidth}{89.40}} & \colorbox{lightred}{\parbox{0.04\textwidth}{89.89}} & \colorbox{lightred}{\parbox{0.04\textwidth}{89.39}} & 91.20 \\
& \text{MaxEnt} &  83.03   & 87.18  & 88.21 & 88.44  & \colorbox{lightred}{\parbox{0.04\textwidth}{92.10}} \\  
& \multicolumn{6}{l}{\cellcolor{lightgray}{Supervised Uncertainty Quantification Method}} \\
&SUQ  & \textbf{85.34}   & \textbf{94.08} & \textbf{93.09} & \textbf{94.07} & \textbf{96.03} \\     \bottomrule
\end{tabular}
\caption{AUC-PR for passage-level Open-ended tasks.}
\label{tb:long:form:tasks:passage}
\end{table}

\section{Study Design}

\noindent\textbf{Dataset.} Our experiments are conducted on public datasets: POPE~\cite{li2023evaluating}, GQA~\cite{hudson2019gqa}, and M-HalDetect~\cite{gunjal2023detecting}. POPE contains three subsets: random, popular, and adversarial. For simplicity, we refer to them as Ran, Pop, and Adv, and we also abbreviate M-HalDetect as M-Hal. In addition, we create a manually annotated, sentence-level dataset, IHAD, by using LLaVA-v1.5-7b to generate concise image descriptions for 500 images from MSCOCO~\cite{lin2014microsoft} using prompt ``Provide a concise description of the given image''. 
Ran, Pop, Adv, and a subset of GQA used in this work evaluate object hallucination as a binary classification task, prompting LVLMs to output ``Yes'' or ``No'', categorizing them as Yes-or-No tasks. In contrast, M-Hal and IHAD involve long text answers based on open-ended questions, making them Open-ended tasks. More detailed dataset information is shown in Appendix~\ref{sec:appendix:dataset}.

\noindent\textbf{Models.}
We evaluate five representative categories of LVLMs: MiniGPT-4v~\cite{zhu2023minigpt}, LLaVA-v1.5-7b~\cite{liu2023visual}, LLaVA-v1.6-vicuna-7b, LLaVA-v1.6-mistral-7b~\cite{liu2023improvedllava} and LLaVA-v1.6-vicuna-13b. These models, featuring different architectures and strong performance, are used as benchmarks in various works~\cite{huang2023opera,zhou2023analyzing}. We only test models of a 7-billion and 13-billion size due to the limited computational resources.

\begin{figure*}[t!]
    \centering
    \begin{subfigure}[b]{0.25\linewidth}
    \includegraphics[width=\textwidth]{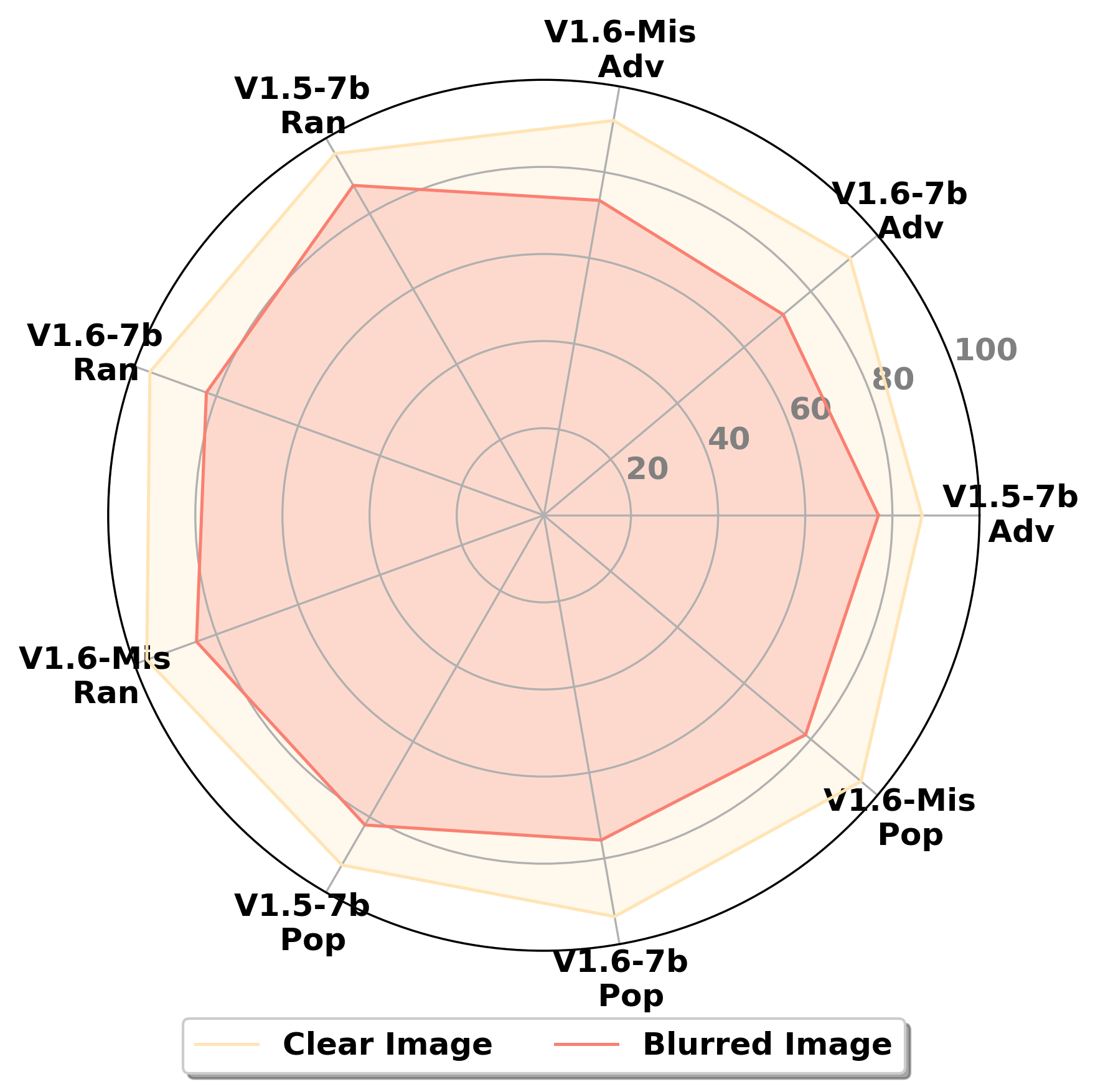}
    \caption{\text{AvgProb}}
    \label{fig:result:logp}
    \end{subfigure}
    \hspace{.00002in}
    \begin{subfigure}[b]{0.25\linewidth}
        \includegraphics[width=\textwidth]{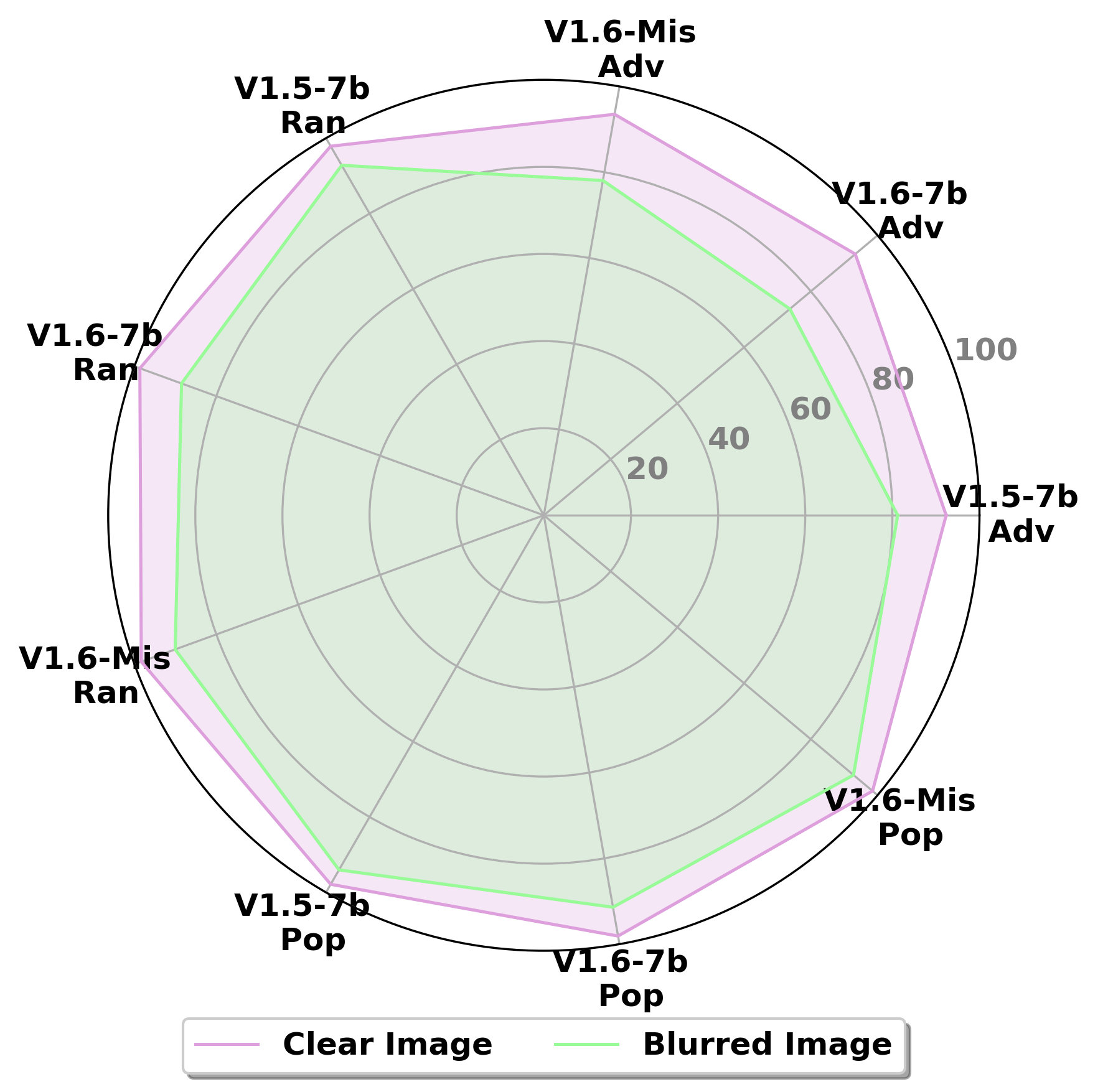}
    \caption{SUQ}
    \label{fig:result:probing}
    \end{subfigure}
    \hspace{.00002in}
    \begin{subfigure}[b]{0.25\linewidth}
        \includegraphics[width=\textwidth]{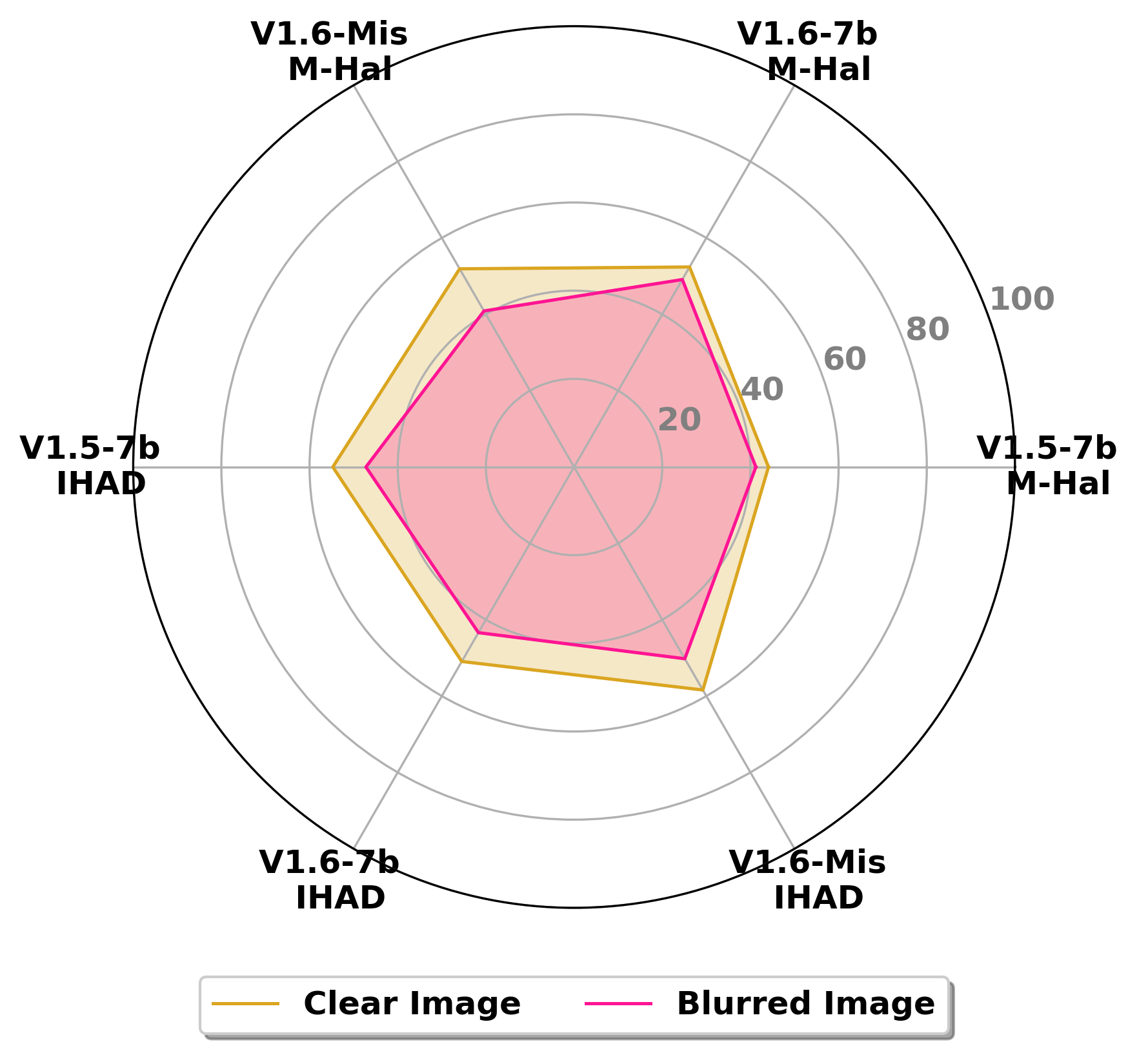}
    \caption{NLI}
    \label{fig:result:nli}
    \end{subfigure}
    \caption{Comparison using images with different clarity.}
    \label{fig:clear:no_clear:comparation}
\end{figure*}

\noindent\textbf{Experimental Setup.}
We combine questions and responses to create inputs for test models. If the response is from this test model, it is considered self-generated input; otherwise, it is regarded as hand-crafted input for this model. Inputs are then processed by test models, from which we extract internal information, including probability, entropy, and embedding states. In this work, all datasets are hand-crafted except for IHAD, which is self-generated data only for LLaVA-v1.5-7b.
Figure~\ref{fig:promt_response.png} shows test model assigns different internal information to two different inputs. Detailed experimental setups are presented in Appendix~\ref{sec:appendix:experimental_setup}.

\section{Experimental Results and Analysis}
The comprehensive results on Open-ended tasks are presented in Tables~\ref{tb:binary:output:tasks}, ~\ref{tb:long:form:tasks:sentence} and~\ref{tb:long:form:tasks:passage}. The numbers in the pink and yellow background, respectively, represent the best results for the uncertainty-based methods and consistency-based methods, while the numbers in bold indicate the overall best results. We utilize AUC-PR~\cite{davis2006relationship}, which stands for the area under the precision-recall curve, to objectively evaluate the effectiveness of each model. The higher the value of AUC-PR, the stronger the ability of this method for hallucination detection.

\subsection{Effectiveness of Various Methods}

\noindent\textbf{Hallucination Detection on Yes-or-No Tasks.} We only compare the uncertainty-based and SUQ methods on Yes-or-No tasks because the model's answers exclusively contain ``Yes'' or ``No''. The random baseline is calculated as the ratio of non-factual examples to the total number of examples. Our results in Table~\ref{tb:binary:output:tasks} demonstrate that \textit{SUQ method outperforms the uncertainty-based methods, exceeding \text{AvgProb} by about $4.5\%$. \text{AvgProb} achieves AUC-PR values nearly double those of the random baseline across all datasets and models, indicating it is an efficient method since it is an unsupervised method.} In addition, the probability $\text{Prob}$ overall performs better than the entropy $\text{Ent}$ in Yes-or-No tasks.
        
\noindent\textbf{Sentence-level Hallucination Detection.}
Based on the data in Table~\ref{tb:long:form:tasks:sentence}, we find that \textit{in most instances, the consistency-based approaches surpasses the uncertainty-based metrics, yet they remain significantly inferior to the SUQ method by approximately 15\%.} Among consistency-based methods, NLI outperforms best while Unigram shows the least effectiveness in most cases. Specifically, on the M-Hal dataset, the NLI performance of LLaVA-v1.6-7b  exceeds the highest performing uncertainty metric, \text{MaxEnt}, by approximately 12\%. BERTScore and QA outperform the uncertain estimation in most setups.

\noindent\textbf{Passage-level Hallucination Detection.}
We calculate the metrics of uncertainty-based methods across all tokens in a passage and take the embedding information of the passage's last token for SUQ. Consistency-based methods are not used to detect non-factual information in passages as the heavy computation. Our results in Table~\ref{tb:long:form:tasks:passage} show that \textit{the SUQ method is still better than the uncertainty-based methods, including probability and entropy methods.  Besides, uncertainty-based methods are weakly better than the random baseline at passage-level.}

\subsection{Robustness and Limitations of SUQ} 
\begin{figure}[t!]
    \centering
    \includegraphics[width=0.8\linewidth]{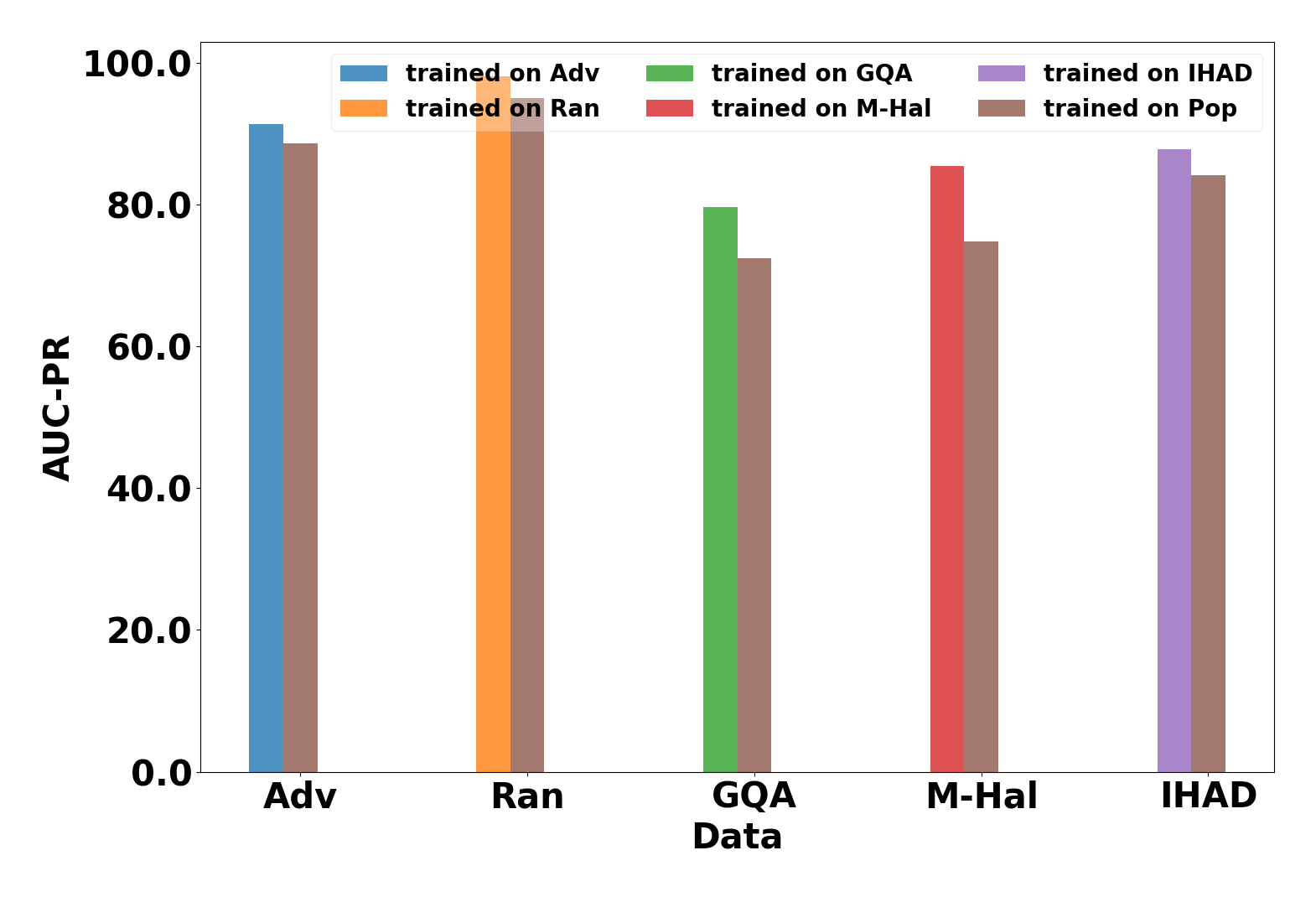}
    \caption{Comparison of using different classifiers.}
    \label{fig:classifiers}
\end{figure}

To investigate the robustness of SUQ methods, we employ two classifiers to assess the impact on each dataset. Each dataset is split into a training set and a test set in a 3:1 ratio. One classifier is trained on the Pop training set, while the other is trained on the training set of each dataset individually. Figure~\ref{fig:classifiers} illustrates that \textit{the classifier trained on Pop exhibits comparable performance to classifiers trained on separated datasets, indicating the robustness of the SUQ method.} 

Although our experiments confirmed the effectiveness of the SUQ method, it was found to be ineffective in detecting subtle, manually crafted hallucinations. For example, the difference between a correct statement, ``There is a red apple and a cute kitten.'' and an incorrect one, ``There is a red orange and a cute kitten.'' has a negligible effect on the model's hidden state at the final position, posing a challenge for SUQ methods to differentiate between them.

\subsection{Impact of Image Clarity}  
\label{sec:appendix:image_clarity}

To explore this question, we employ the Gaussian blur technique, which smooths the image by assigning a weighted average to the surrounding pixels of each pixel, resulting in an obscure picture. We set $\text{radius}=10$ in our experiments. Figure~\ref{fig:clear:no_clear:comparation} shows the effectiveness of \text{AvgProb}, SUQ, and NLI in blurred images are significantly reduced compared to clear images. The phenomenon suggests that \textit{enhancing image clarity is a crucial strategy for boosting the effectiveness of hallucination detection methods.}

\subsection{Impact of Data Source } 
\label{sec:data_sources}

Reference-free methods, particularly uncertain-based and SUQ methods, rely on the internal information of models. A natural question is whether these methods are affected by data sources. M-Hal and IHAD are hand-crafted data and self-generated data, respectively, for LLaVA-v1.5-7b. Therefore, we compare various detection methods on M-Hal and IHAD based on LLaVA-v1.5-7b. Specifically, we calculate $\Delta \text{AUC-PR}$ 
\begin{equation*}
  \Delta \text{AUC-PR}_{i,j} = \text{AUC-PR}_{i,j} - \text{AUC-PR}_{i,\text{baseline}},
\end{equation*}
where $i$ refers to M-Hal and IHAD, $j$ denotes different approaches.

\begin{figure}[t!]
    \centering
    \includegraphics[width=0.87\linewidth]{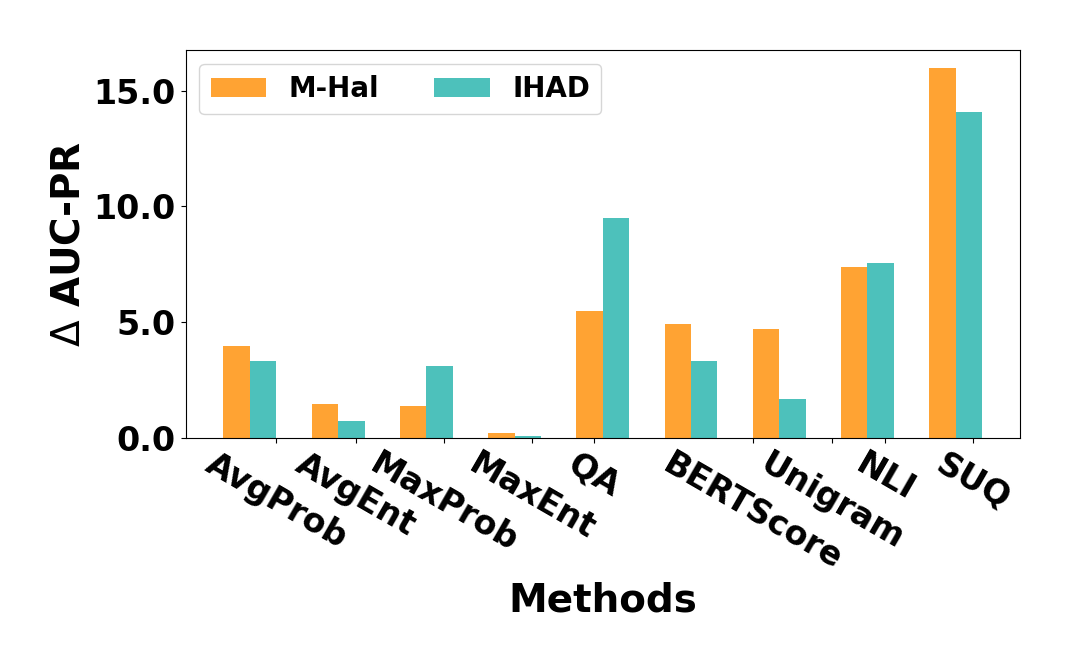}
    \caption{Comparison of using different data sources.}
    \label{fig:data_sources}
\end{figure}

Figure~\ref{fig:data_sources} shows that in uncertainty-based approaches, some metrics outperform in M-Hal while others excel in IHAD, a trend also observed in consistency-based techniques. This phenomenon illustrates that \textit{the reference-free hallucination detection methods are insensitive to the data source.} 

\noindent\textbf{Others.} We have also explored the impact of unimportant tokens for uncertainty-based methods and the optimal hidden layers for SUQ. They are discussed respectively in Appendices~\ref{sec:appendix:unimportant_token} and~\ref{sec:appendix:probing_methods}.

\section{Conclusion}
This paper systematically studies reference-free approaches for detecting the hallucination of LVLMs. Our results indicate that the SUQ method performs best as a supervised method, but it relies on sufficient training data, which could limit its applicability in certain scenarios. Consistency-based methods outperform uncertainty-based methods; however, they require more computation and are more suitable for black-box models. Uncertainty-based methods are particularly well-suited for tasks similar to Yes-or-No questions. 

\section*{Limitations}
Three major limitations are identified in this work. First, we did not compare with reference-based methods. The performance of reference-based methods depends on the performance of external models and is specific to certain tasks. In this article, we primarily aim to explore the potential of reference-free methods, which have a broader range of applications. Second, we did not classify types of hallucinations; limited by resources, even with fine-grained annotations, we only marked whether sentences contained hallucinations without distinguishing types such as object existence errors, attribute errors, text recognition errors, and fact-conflicting errors. This might be where SUQ methods have an advantage. Third, we only studied white-box models. Recent research indicates that white-box models can be proxies for black-box models to calculate their confidence regarding specific issues. Unigram method is one such proxy model approach. In the future, we could use more complex models, closer in architecture to the target models, as proxies to explore their potential in hallucination detection.

\section*{Ethics and Broader Impact}
We sampled a portion of the data from existing datasets for our experiments, which may affect the accuracy of some of our conclusions.

\bibliography{main-short-final}

\appendix

\section{Reference-free Hallucination Detection Methods}
We collect uncertainty-based and consistency-based methods proposed in~\cite{manakul2023selfcheckgpt} to test the effectiveness on LVLMs.

\subsection{Uncertainty-based methods}
\label{sec:appendix:uncertainty-based-methods}

To aggregate the uncertainty information obtained at the token level, we employ four metrics to aggregate token-level uncertainty into sentence level. In particular, a sentence-level uncertainty score can be obtained by taking either the maximum or average of the negative loglikelihood $-\log p_{ij}$ in a sentence:
\begin{align}
    \text{MaxProb}(i) &= \max_{j} (-\log p_{ij}),
    \\
    \text{AvgProb}(i) &= -\frac{1}{J_i} \sum_{j = 1}^{J_i} \log p_{ij},
\end{align}
where $p_{ij}$ is the probability of a token at a position $j$ in the sentence $i$ and $J_i$ is the total number of tokens in the considered sentence. Additionally, one can also replace the negative loglikelihood $-\log p_{ij}$ with the entropy  $\mathcal{H}_{ij}$:
\begin{align}
    \text{MaxEnt}(i) &= \max_{j} \mathcal{H}_{ij},
    \\
    \text{AvgEnt}(i) &= \frac{1}{J_i} \sum_{j = 1}^{J_i} \mathcal{H}_{ij},
\end{align}
where $H_{ij}$ is the entropy of the token distribution for the $j$-th token in the sentence $i$.

\subsection{Consistency-based Methods}
\label{sec:appendix:consistency-based-methods}

\noindent\textbf{BERTScore} denoted as $\mathcal{S}_{BERT}$, finds the average BERTScore of the considered sentence with the most similar sentence from each drawn sample. Let $\mathcal{B}(.,.)$ denote the BERTScore between two sentences. $\mathcal{S}_{BERT}$ is computed by $\mathcal{B}(.,.)$ as:
\begin{equation}
  \mathcal{S}_{\text{BERT}}(i) = 1 - \frac{1}{N}\sum_{n=1}^{N} \max\limits_{k} \mathcal{B}(r_i,s_k^n),
\end{equation}
where $r_i$ represents the $i$-th sentence in main response $R$, $s_k^n$ represents the $k$-th sentence in the $n$-th sample $S^n$ and $N$ is the total number of sampled sentences.

\noindent\textbf{Question Answering (QA).} The multiple-choice question answering generation (MQAG) evaluates consistency by creating multiple-choice questions from the primary generated response and answering them using other sampled responses as reference. MQAG consists of two stages: question generation $G$ and question answering $A$. For the sentence $r_i$ in the response $R$, we draw questions $q$ and options $\mathbf{o}$:
\begin{equation}
  q, \mathbf{o} \sim P_G(q, \mathbf{o} \mid r_i, R).
\end{equation}
The answering stage $A$ selects the answers:
\begin{align}
  a_R &= \mathop{\arg\max}_{k}[P_A(o_k \mid q, R, \mathbf{o})],
  \\
  a_{S^n} &= \mathop{\arg\max}_{k}[P_A(o_k \mid q, S^n, \mathbf{o})].
\end{align}
We compare whether $a_R$ is equal to $a_{S^n}$ for each sample in $\{ S^1, \dots, S^N \}$, yielding $N_m$ matches and $N_n$ not-matches. A simple inconsistency score for the $i$-th sentence and question $q$ based on the match/not-match counts is defined: $S_{\text{QA}}(i, q) = \frac{N_n}{N_n + N_m}$. \citet{manakul2023selfcheckgpt} modify the inconsistency score $S_{\text{QA}}(i, q)$ by incorporating the answerability of generated questions.
Finally, $\mathcal{S}_{\text{QA}}(i)$ is the average of inconsistency scores across $q$,
\begin{equation}
  \mathcal{S}_{\text{QA}}(i) = E_q [S_{\text{QA}}(i, q)].
\end{equation}

\noindent\textbf{Unigram.} The concept behind Unigram is to develop a new model that approximates the LVLMs by samples $\{S^1, \dots, S^N\}$ and get the LVLM's token probabilities using this model. As $N$ increases, the new model gets closer to LVLMs. Due to time and cost constraints, we just train a simple $n$-gram model using the samples $\{S^1, \dots, S^N\}$ as well as the main response $R$. We then compare the average and maximum of the negative probabilities of the sentence in response $R$ using the following equations:
\begin{align}
  \mathcal{S}^{\text{Avg}}_{\text{n-gram}}(i) &= -\frac{1}{J_i} \sum_{j = 1}^{J_i} \log \hat{p}_{ij},
  \\
  \mathcal{S}^{\text{Max}}_{\text{n-gram}}(i) &= \max_{j}(-\log \hat{p}_{ij}),
\end{align}
where $\hat{p}_{ij}$ is the probability of a token at position $j$ of a sentence $i$.

\noindent\textbf{Natural Language Inference (NLI)} determines whether a hypothesis follows a premise, classified into either entailment/neutral/contradiction. In this work,  we use DeBERTa-v3-large~\cite{DBLP:conf/iclr/HeGC23} fine-tuned to MNLI as the NLI model. The input for NLI classifiers is typically the premise concatenated to the hypothesis, which for NLI  is the sampled passage $S^n$ concatenated to the sentence to be assessed $r_i$ in the response $R$.
Only the logits associated with the ‘entailment’ and ‘contradiction’ classes are considered,
\begin{equation*}
  P(contradict \mid r_i, S^n) = \frac{\exp\bigl(z_e^{i,n}\bigr)}{\exp\bigl(z_e^{i,n}\bigr) + \exp\bigl(z_c^{i,n}\bigr)},
\end{equation*}
where $z_e^{i,n} = z_e(r_i, S^n)$ and $z_c^{i,n} = z_c(r_i, S^n)$ are the logits of the ‘entailment’ and ‘contradiction’ classes. NLI score for sentence $r_i$ on samples $\{S^1, \dots, S^N\}$ is then defined as,
\begin{equation}
  \mathcal{S}_{\text{NLI}}(i) = \frac{1}{N} \sum_{n=1}^{N}P(contradict \mid r_i, S^n).
\end{equation}

\section{Dataset and Implementation}
\label{sec:appendix:dataset_implementation}

\subsection{Dataset}
\label{sec:appendix:dataset}

\noindent\textbf{POPE}~\cite{li2023evaluating} formulates the evaluation of object hallucination as a binary classification task that prompts LVLMs to output ``Yes'' or ``No'', e,g., ``Is there a chair in the figure?'' POPE adopts three sampling settings to construct negative samples: random, popular, and adversarial. The random setting selects non-present objects at random. In contrast, the popular setting chooses from a list of frequently occurring but absent objects, and the adversarial method selects based on common co-occurrence in contexts despite the absence in the target image.

\noindent\textbf{GQA}~\cite{hudson2019gqa} is proposed to detect hallucinations, including existence, attribute, and relation. It contains not only Yes-or-No problems but also Open-ended problems. We randomly 9050  some binary problems for 397 images used in our experiments.

\noindent\textbf{M-HalDetect}~\cite{gunjal2023detecting} consists of 16k fine-grained annotations on visual question-answering examples, including object hallucination, descriptions, and relationships. We utilize about 3000 questions on 764 figures for our Open-ended tasks, averaging 4 sentences per response.

\noindent\textbf{IHAD} severs as the evaluation of description for Open-ended tasks.  It is a sentence-level dataset created by prompting LLaVA-V1.5-7b to generate concise image descriptions for 500 images from MSCOCO~\cite{lin2014microsoft}.
On average, each response contains 5 sentences. The creation of IHAD aims to investigate if the data source impacts the internal information of LVLMs and further impacts hallucination detection methods, which rely on this internal information.

\begin{table}[t!]
\small
\centering
\begin{tabular}{m{0.05cm}<{\centering} m{1.5cm}<{\raggedright} m{0.8cm}<{\centering} m{0.8cm}<{\centering} m{0.8cm}<{\centering} m{0.8cm}<{\centering}}
\toprule
   & \multirow{2}{*}{\begin{tabular}[c]{@{}l@{}} {Method} \end{tabular}}     & Mini & LLaVA & LLaVA & LLaVA \\
    &           & GPT-v4     & v1.5-7b         &v1.6-vicuna-7b       &v1.6-mistral-7b \\
\midrule
\multirow{5}{*}{\begin{tabular}[c]{@{}l@{}} \rotatebox{90}{Adv} \end{tabular}} 
&Baseline & 48.73     & 48.73   & 48.73     & 48.73           \\ 
& \multicolumn{5}{l}{\cellcolor{lightgray}{Uncertainty-based Methods}}  \\
& \text{AvgProb} & \colorbox{lightred}{\parbox{0.04\textwidth}{55.87}} & \colorbox{lightred}{\parbox{0.04\textwidth}{58.53}} & \colorbox{lightred}{\parbox{0.04\textwidth}{54.27}}  & \colorbox{lightred}{\parbox{0.04\textwidth}{81.70}}   \\
& \text{AvgEnt}    & 51.74 &50.87 & 48.82 & 49.41    \\           
& \text{MaxProb} & 32.28 &36.48&  32.11 & 41.78    \\
& \text{MaxEnt}    & 49.03 &50.60& 48.35   & 49.25   \\ 
\midrule
\multirow{5}{*}{\begin{tabular}[c]{@{}l@{}} \rotatebox{90}{Ran} \end{tabular}}   
&Baseline    & 48.73     & 48.73   &   48.73 & 48.73            \\
& \multicolumn{5}{l}{\cellcolor{lightgray}{Uncertainty-based Methods}}  \\
& \text{AvgProb} & \colorbox{lightred}{\parbox{0.04\textwidth}{54.57}} & \colorbox{lightred}{\parbox{0.04\textwidth}{51.95}}&  47.35 &   \colorbox{lightred}{\parbox{0.04\textwidth}{85.10}}  \\
& \text{AvgEnt} & 50.49 & 51.13 & \colorbox{lightred}{\parbox{0.04\textwidth}{53.51}} &   50.80  \\           
& \text{MaxProb} & 39.80 &33.77 &   30.68 &  41.30   \\
& \text{MaxEnt} & 48.37 & 50.79&   51.43 &   50.75  \\           
\midrule
\multirow{5}{*}{\begin{tabular}[c]{@{}l@{}} \rotatebox{90}{Pop} \end{tabular}} 
&Baseline    & 48.73     & 48.73  &  48.73  & 48.73              \\
& \multicolumn{5}{l}{\cellcolor{lightgray}{Uncertainty-based Methods}}   \\
& \text{AvgProb} & \colorbox{lightred}{\parbox{0.04\textwidth}{55.87}} & \colorbox{lightred}{\parbox{0.04\textwidth}{57.65}} & 51.52 & \colorbox{lightred}{\parbox{0.04\textwidth}{81.74}} \\
& \text{AvgEnt}   & 51.74 & 50.59 &  \colorbox{lightred}{\parbox{0.04\textwidth}{51.83}} & 51.92 \\        
& \text{MaxProb} & 32.28 & 35.90 & 31.41 & 40.97  \\
& \text{MaxEnt}  & 49.03 &50.30 & 50.95 &  51.73  \\
\midrule
\multirow{5}{*}{\begin{tabular}[c]{@{}l@{}} \rotatebox{90}{GQA} \end{tabular}}
&Baseline    & 48.47     & 48.47   &  48.47  & 48.47         \\
& \multicolumn{5}{l}{\cellcolor{lightgray}{Uncertainty-based Methods}}  \\
& \text{AvgProb} & \colorbox{lightred}{\parbox{0.04\textwidth}{55.72}} &\colorbox{lightred}{\parbox{0.04\textwidth}{56.56}} & \colorbox{lightred}{\parbox{0.04\textwidth}{57.93}} & \colorbox{lightred}{\parbox{0.04\textwidth}{65.06}} \\
& \text{AvgEnt}    & 48.96 &50.12& 49.55& 48.38 \\         
& \text{MaxProb} & 33.20 &39.64& 37.27& 46.25\\
& \text{MaxEnt}    & 48.16 &49.98& 49.15& 48.39\\
\bottomrule
\end{tabular}
\caption{Yes-or-No tasks with period. The terms \colorbox{lightred}{\parbox{0.03\textwidth}{top}} denote the highest AUC-PR values for uncertainty-based methods.}
\label{tb:yes:or:no:tasks:with:period:all}
\end{table}

\subsection{Experimental Setup}
\label{sec:appendix:experimental_setup}

In our experiments, we merge questions and responses from humans or other models to create hand-crafted input. This input is then processed by the evaluation model, from which we extract internal information, including probability, entropy, and embedding states. The evaluation model, capable of recognizing hallucinations~\cite{duan2024llms}, can assign reasonable internal information to responses not generated by itself. As shown in Figure~\ref{fig:promt_response.png}, when generating the ``cat'' token, the evaluation model assigns probabilities to all tokens and selects the higher probability token for output. Specifically, the probability assigned to ``cat'' in self-generated input is 0.9, while in hand-crafted input, ``dog'' receives a probability of 0.1. POPE, GQA, and M-Hal are hand-crafted data for all four evaluation models used in this work. IHAD's response, however, is generated by LLaVA-v1.5-7b, making IHAD self-generated data specifically for LLaVA-v1.5-7b. For the other three models, IHAD is considered hand-crafted data. 

In the Yes-or-No tasks, we simplify the original responses to ``Yes'' for positive answers (e.g., ``Yes, there is a cute cat.'') and ``No'' for all others. To balance positive and negative samples, we create two entries for each original datum, one with a ``Yes'' response and the other with a ``No'' response. Thus, the original datasets Adv, Ran, and Pop, each containing 3000 questions for 500 images, are expanded to 6000 questions per set for the same images. Similarly, GQA is increased to 9050 questions for 397 figures. Regarding Open-ended tasks, each response contains an average of 4 sentences in M-HalDetect, while 5 sentences in IHAD.

In the consistency-based method, we set the temperature to 0.0 to obtain the main resource and use standard beam search decoding. In addition, we set the temperature to 1.0 for the stochastically generated samples and generated $N=10$ samples. 

The classifier used in the SUQ method employs a feedforward neural network featuring three hidden layers with decreasing numbers of hidden units (256, 128, 64), all utilizing ReLU activations. The final layer is a sigmoid output. We use the Adam optimizer. We do not fine-tune any of these hyper-parameters for this task. The classifier is trained for 20 epochs. We use about two-thirds of a dataset to train a classifier based on a specific model and then test its accuracy on the other third on the same model. The trained well classifier is required to determine which sentences the LVLM ``believes'' are true and which it ``believes'' are false.

\section{Experimental Results}

\subsection{Impact of Unimportant Tokens for Uncertainty-based Methods}
\label{sec:appendix:unimportant_token}
It is observed that the performance of uncertainty methods on the Open-ended tasks is not as good as on the Yes-or-No tasks. To explore the reasons, we append a period, resulting in ``Yes.'' or ``No.'' in the Yes-or-No task. In this case, probability $p$ and entropy $\mathcal{H}$ of two tokens need to be considered. 
Table~\ref{tb:yes:or:no:tasks:with:period:all} shows the detailed performance for Yes-or-No tasks with period using different methods. Specifically, for LLaVA-v1.6-mistral, the \text{AvgProb} scores decline to 81.70 from 92.04 for Adv, to 85.10 from 96.99 for Ran, to 81.74 from 95.06 for Pop, and to 65.06 from 79.15 for GQA datasets, indicating the performance of \text{AvgProb} is severely affected by unimportant tokens. The other three metrics, $\text{AvgEnt}$, \text{MaxProb}, and \text{MaxEnt}, nearly equal to the baseline and play no role in hallucination detection.

\subsection{Optimal Hidden Layers for Effective SUQ Method}
\label{sec:appendix:probing_methods}

\begin{figure}[hbt!]
    \centering
    \begin{subfigure}[b]{0.235\textwidth}
    \includegraphics[width=\textwidth]{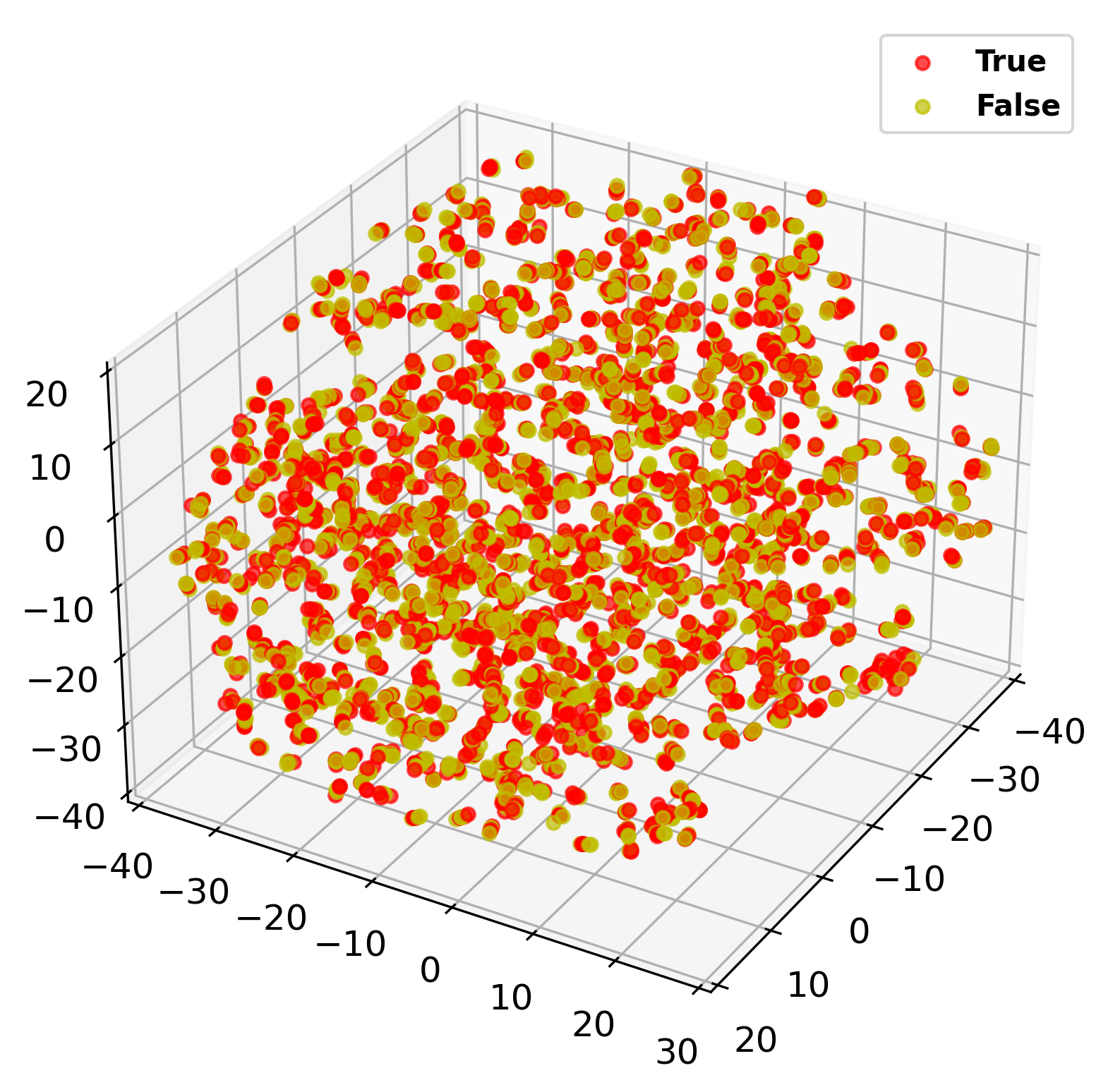}
    \caption{Fifth Layer}
    \label{fig:layer:4}
    \end{subfigure}
    \begin{subfigure}[b]{0.235\textwidth}
        \includegraphics[width=\textwidth]{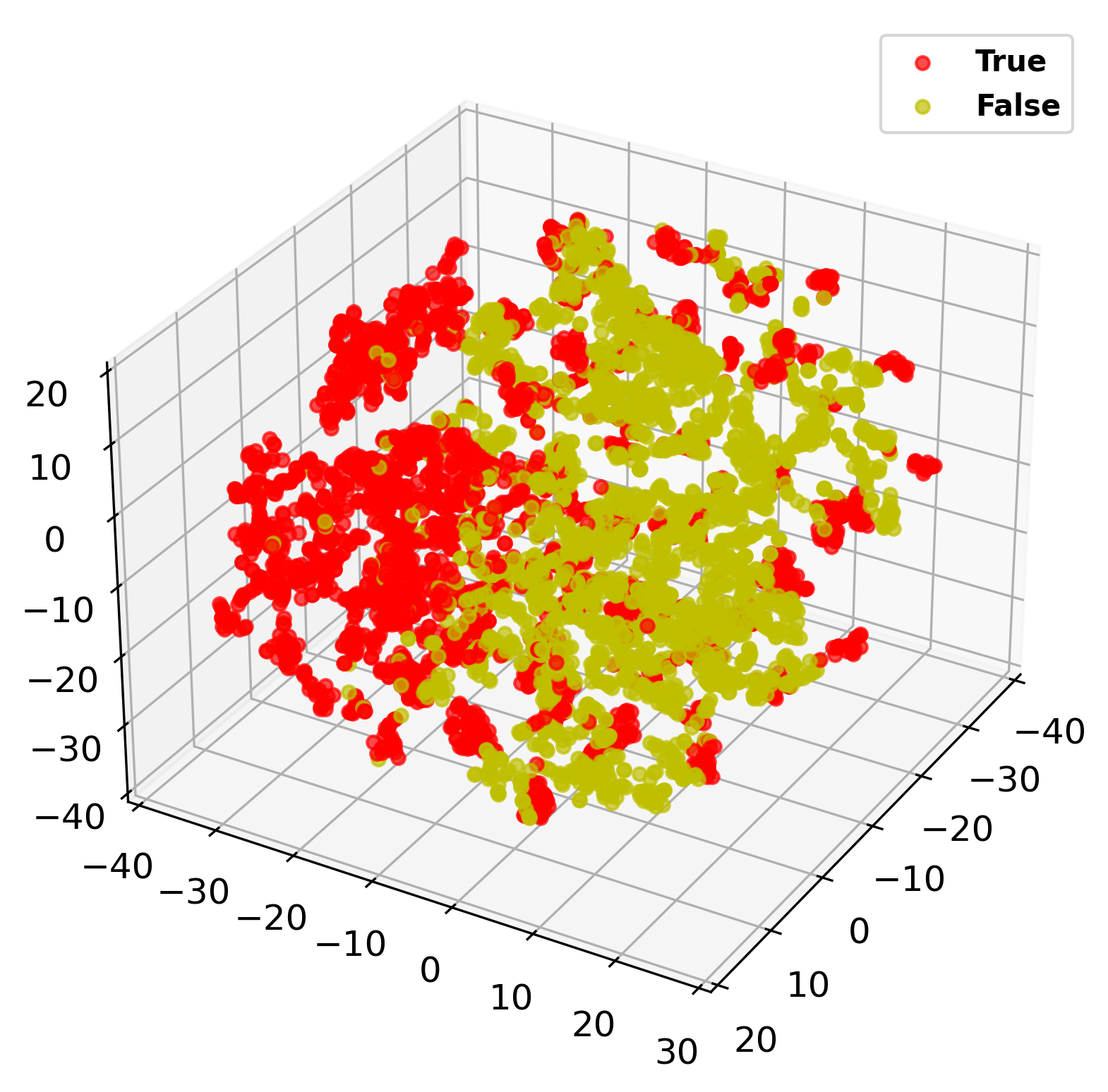}
    \caption{Last Layer}
    \label{fig:layer:32}
    \end{subfigure}
    \caption{3D TSNE plot of language embeddings computed from the fifth and last layers of LLaVA-v1.6-7b on popular task.}
    \label{fig:hidden:layers}
\end{figure}

To explore which hidden layer information should be used, we perform 3D TSNE visualizations of the embedding states of LLaVA at the fifth and last layers for popular tasks, as shown in Figure~\ref{fig:hidden:layers}. These visualizations show that at the last layer, there are distinct separations between the representations of different labels. Conversely, at the fifth layer, representations of different labels are relatively close and almost blend together, indicating that the shallow layers primarily preserve language-agnostic features. Therefore, we use the last hidden layer to explore the SUQ method in our experiments.

\subsection{The comparison of clear and blurred images.}
\label{sec:appendix:different_figures}

\begin{figure}[hbt!]
    \centering
    \begin{subfigure}[b]{0.23\textwidth}
    \includegraphics[width=\textwidth]{images/clear_figure.jpg}
    \caption{Clear Image}
    \label{fig:clear:image}
    \end{subfigure}
    \hspace{.00002in}
    \begin{subfigure}[b]{0.23\textwidth}
        \includegraphics[width=\textwidth]{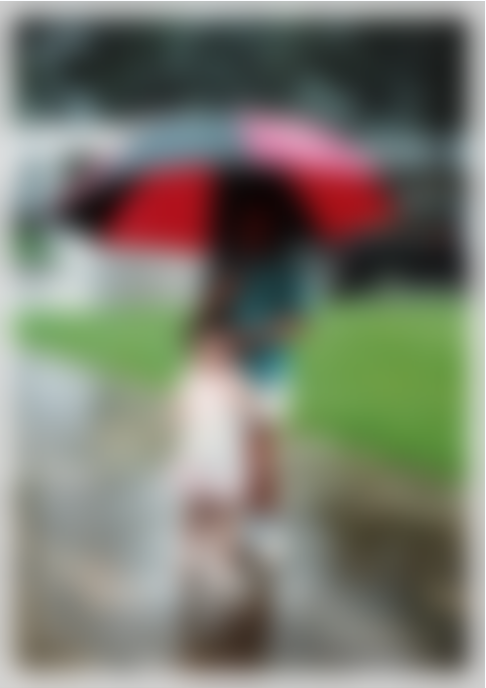}
    \caption{Blurred Image}
    \label{fig:blurred:image}
    \end{subfigure}
    \caption{The comparison of clear and blurred images.}
    \label{fig:clear:no_clear:figures}
\end{figure}

The comparison of clear and blurred images used in \ref{sec:appendix:image_clarity} is illustrated in Figure~\ref{fig:clear:no_clear:figures}.

\end{document}